\documentclass{article}
\usepackage[backref=page]{hyperref}
\usepackage[utf8]{inputenc}
\usepackage{amsmath,amsthm,amssymb}
\usepackage{fullpage}
\usepackage{url}
\usepackage{natbib}
\usepackage{algorithm}
\usepackage{algorithmic}

\newcommand{\be}{\boldsymbol{e}}
\newcommand{\bg}{\boldsymbol{g}}

\newcommand{\bx}{\boldsymbol{x}}
\newcommand{\bu}{\boldsymbol{u}}

\newcommand{\bpi}{\boldsymbol{\pi}}

\newcommand{\field}[1]{\mathbb{#1}}

\newcommand{\R}{\field{R}}

\DeclareMathOperator{\KL}{KL}

\newcommand{\indicator}{\mathbf{1}}

\title{A Note on How to Remove the $\ln\ln T$ Term from the Squint Bound}
\author{Francesco Orabona\\
King Abdullah University of Science and Technology (KAUST)\\
Thuwal, 23955-6900, Kingdom of Saudi Arabia\\
\url{francesco@orabona.com}
}

\begin{document}
\maketitle

\begin{abstract}
In \citet{OrabonaP16}, we introduced the \emph{shifted KT potentials}, to remove the $\ln \ln T$ factor in the parameter-free learning with expert bound.
In this short technical note, I show that this is equivalent to changing the prior in the Krichevsky--Trofimov algorithm. Then, I show how to use the same idea to remove the $\ln \ln T$ factor in the data-independent bound for the Squint algorithm.
\end{abstract}

\section{Background}
Parameter-free online learning algorithms, by definition, are those that satisfies the optimal regret bound, up to polylogarithmic multiplicative factors, with respect to $T$ and to all the comparators in the feasible set $\mathcal{V} \subseteq\R^d$~\citep{Orabona19}.

An example of this class of algorithms is Squint~\citep{KoolenVE15} that achieves\footnote{This particular bound appears in \url{http://blog.wouterkoolen.info/Squint_PAC/post.html}.} a regret upper bound of
\begin{align*}
\sum_{t=1}^T \langle \bg_t, \bx_t - \bu\rangle
&\leq \sqrt{2 V_T(\bu)}
\left[ 1 + \sqrt{2 \left( \KL(\bu;\bpi) + \ln\left(\frac{1}{2} + \ln (T+1) \right) \right)}\right]\\
&\quad + 1 + 5\big(\KL(\bu;\bpi) + \ln \left( 1 + 2\ln (T+1) \right)
\big), \quad \forall \bu \in \Delta^{d-1},
\end{align*}
where $\bpi$ is a given prior distribution over the $d$ experts, $\KL$ is the Kullback--Leibler divergence, $V_T(\bu)=\sum_{t=1}^T \sum_{i=1}^d u_i (\langle\bg_t,\bx_t-\be_i\rangle)^2$, and $g_{t,i}\in [0,1]$ for all $i=1,\dots, d$ and $t=1, \dots, T$.

Under the same assumptions, \citep{OrabonaP16} (see also \citet{ChernovV10}) achieved the bound
\[
\sum_{t=1}^T \langle \bg_t, \bx_t - \bu\rangle
\leq \sqrt{3 T( \KL(\bu;\bpi) + 3)}, \quad \forall \bu \in \Delta^{d-1},
\]
that has a better dependence on $T$ in the worst-case ($\mathcal{O}(\sqrt{T})$ versus $\mathcal{O}(\sqrt{T \ln \ln T})$ when $T\to\infty$), but it is data-independent, in the sense that it does not depend on the $\bg_t$ anymore. I want to stress that this second bound is not always better than the one of Squint above.
Yet, a natural question is if it is possible to change the Squint algorithm to achieve the same bound, removing the $\ln \ln T$ term.

\section{Main Result}
We will follow the construction in \citet{OrabonaP16} and \citet{Orabona19}, that reduces the learning-with-expert problem to a gambling game on a continuous coin. Then, the final bound depends on the regret of the gambling algorithms. Given that this is a short note, we refer the reader to the two works above for the details of the construction, and we will directly focus on the final result. Note that we will \emph{not} follow the construction in the original Squint paper that integrates only over positive betting fraction because it complicates the proof.

Consider the prior
\begin{equation}
\label{eq:prior}
F(\beta)
=\frac{(1+\beta)^z (1-\beta)^z}{\int_{-1}^{1} \! (1+\beta)^z(1-\beta)^z\,\mathrm{d}\beta}
= \frac{(1+\beta)^z (1-\beta)^z}{2^{2z+1}
\frac{\Gamma(z+1)^2}{\Gamma(2z+2)}},
\end{equation}
where $\beta \in [-1,1]$.

Then, the wealth of a mixture betting algorithm on coin outcomes $c_t \in \{-1,1\}$ using the prior $F$ is
\[
\int_{-1}^{1}\! (1+\beta)^{a}(1-\beta)^{b} F(\beta) \, \mathrm{d}\beta
= 2^T\frac{\Gamma(a+z+1)\Gamma(b+z+1)\Gamma(2z+2)}{\Gamma(T+2z+2)\Gamma(z+1)^2},
\]
where $a=\sum_{t=1}^T \indicator_{\{c_t=1\}}$ and $b=\sum_{t=1}^T \indicator_{\{c_t=-1\}}$.

The first observation is that the expression above is exactly the ``shifted KT'' potential introduced in \citet{OrabonaP16} to get rid of the $\ln \ln$ term in the bound, even if at that time we did not know how to obtain such potential through a prior.
In \citet{OrabonaP16} we chose $z=\frac{\delta-1}{2}$ and $\delta$ of the order of $T$. This corresponds to a prior peaked at 0, that makes sense when we think that we are aiming at obtaining a strategy that loses only a fraction of its money, hence with conservative bets. Such a choice guarantees that the betting procedure will never lose more than a fixed fraction of the money independent from $T$, on $T$ rounds.

Now, consider the Squint potential~\citep{KoolenVE15}, modified to integrate over positive and negative betting fractions:
\[
W_t(x)
=\int_{-1/2}^{1/2} \! \exp\left(\beta x - \beta^2 \sum_{i=1}^{t} c^2_i\right) F(\beta)\, \mathrm{d}\beta~.
\]
Note that we integrate over $[-1/2,1/2]$ instead of $[-1,1]$ of the Krichevsky--Trofimov strategy.
This is a valid betting potential for any prior $F$, following the definition in \citet{OrabonaP16}. In particular, it gives rise to a betting strategy that bets a signed fraction of the current wealth equal to
\[
\beta_t
= \frac{\int_{-1/2}^{1/2} \! \beta \exp\left(\beta \sum_{i=1}^{t-1} c_i - \beta^2 \sum_{i=1}^{t-1} c^2_i\right) F(\beta)\, \mathrm{d}\beta}{\int_{-1/2}^{1/2} \! \exp\left(\beta \sum_{i=1}^{t-1} c_i - \beta^2 \sum_{i=1}^{t-1} c^2_i\right) F(\beta)\, \mathrm{d}\beta}
,
\]
which guarantees that the wealth at the end of round $t$ is at least $W_t(\sum_{i=1}^t c_i)$, for all $t$.

If we do not care about obtaining data-dependent bounds, we can use a worse betting potential, by lower bounding $W_t$:
\[
\widetilde{W}_t(x)=
\int_{-1/2}^{1/2} \! \exp\left(\beta x - \beta^2 t\right) F(\beta)\, \mathrm{d}\beta,
\]
where we used the fact that $|c_t|\leq 1$.
Now, we have to choose the prior for this betting potential.

It is important to remember that the Squint potential comes from the lower bound $1+\beta c_t \geq \exp(\beta c_t - \beta^2 c_t^2)$ for $|\beta c_t|\leq 1/2$.
So, let us approximate of the prior in \eqref{eq:prior}, using the same lower bound, to have the same shape of the Squint potential:
\[
(1+\beta)^z (1-\beta)^z
\geq \exp(z\beta - z \beta^2 - z \beta - z \beta^2)
= \exp( -2 z \beta^2)~.
\]
Hence, it has a Gaussian shape. So, now it is natural to choose a Gaussian prior truncated in $[-1/2,1/2]$ in $\widetilde{W}_t$ and see if it removes the $\ln \ln$ term in the data-independent bound of Squint. So, we have
\begin{align*}
\widetilde{W}_T(x)
&=\frac{\int_{-1/2}^{1/2} \! \exp\left(\beta x - \beta^2 T\right) \exp\left(-\frac{\beta^2}{2\sigma^2}\right)\, \mathrm{d}\beta}{\int_{-1/2}^{1/2} \!  \exp\left(-\frac{\beta^2}{2\sigma^2}\right)\, \mathrm{d}\beta}\\
& = \frac{
\exp\left(\frac{x^2}{4\lambda}\right)
}{
2\sqrt{2}\,\sigma\sqrt{\lambda}\, \operatorname{erf}\left(\frac{1}{2\sqrt{2}\sigma}\right) }\left[\operatorname{erf}\left(\frac{\sqrt{\lambda}}{2}-\frac{x}{2\sqrt{\lambda}}\right)+\operatorname{erf}\left(\frac{\sqrt{\lambda}}{2}+\frac{x}{2\sqrt{\lambda}}\right)\right],
\end{align*}
where $\lambda=T+\frac{1}{2\sigma^2}$.

Set, for example, $\sigma^2=\frac{1}{2T}$, to have $\lambda=2T$ and $\sigma \sqrt{\lambda}=1$. So, we have
\begin{align*}
\widetilde{W}_T(x)
&=\frac{\int_{-1/2}^{1/2} \! \exp\left(\beta x - \beta^2 T\right) \exp\left(-\frac{\beta^2}{2\sigma^2}\right)\, \mathrm{d}\beta}{\int_{-1/2}^{1/2} \!  \exp\left(-\frac{\beta^2}{2\sigma^2}\right)\, \mathrm{d}\beta}\\
&= \frac{1}{2\sqrt{2} \operatorname{erf}\left(\frac{\sqrt{T}}{2}\right)}
\exp\left(\frac{x^2}{8T}\right)\left[\operatorname{erf}\left(\frac{\sqrt{2T}}{2}-\frac{x}{2\sqrt{2T}}\right)+\operatorname{erf}\left(\frac{\sqrt{2T}}{2}+\frac{x}{2\sqrt{2T}}\right)\right]~.
\end{align*}
We will now evaluate $\widetilde{W}_T$ in $\sum_{t=1}^T c_t$ to derive the wealth lower bound after $T$ rounds. Using the fact that $|\sum_{t=1}^T c_t|\leq T$, we can lower bound all the terms with the erf:
\begin{align*}
\widetilde{W}_T\left(\sum_{t=1}^T c_t\right)
&\geq \frac{1}{2\sqrt{2} \operatorname{erf}\left(\frac{\sqrt{T}}{2}\right)}
\left[\operatorname{erf}\left(
\frac{\sqrt{2T}}{2}-\frac{\sum_{t=1}^T c_t}{2\sqrt{2T}} \right) + \operatorname{erf}\left( \frac{\sqrt{2T}}{2}+\frac{\sum_{t=1}^T c_t}{2\sqrt{2T}} \right)\right]\\
&\geq \frac{\operatorname{erf}\left(\sqrt{T}\left(\frac{\sqrt{2}}{2}-\frac{1}{2\sqrt{2}}\right)\right)}{\sqrt{2} \operatorname{erf}\left(\frac{\sqrt{T}}{2}\right)}
> \frac{1}{2}~.
\end{align*}

Hence, overall we have
\[
\widetilde{W}_T(s)
\geq \frac{1}{2} \exp\left(\frac{s^2}{8T}\right)~.
\]
Finally, it is now enough to use the reduction from betting to learning with experts~\citep{OrabonaP16,Orabona19} to obtain
\[
\sum_{t=1}^T \langle \bg_t, \bx_t - \bu\rangle
\leq \sqrt{8 T( \KL(\bu;\bpi) + \ln 2)}, \quad \forall \bu \in \Delta^{d-1},
\]
where $\bpi$ is the prior distribution over the experts. We can also get rid of the knowledge of $T$ by using a simple doubling trick.

\section{History Bits}

Effectively, this is a result from 10 years ago, but it was never written anywhere.
In fact, when in 2016 I described to Wouter Koolen and Tim van Erven the shifted KT potentials to remove the $\ln \ln T$ term, they immediately replied ``essentially you are putting more mass around 0''. That made a lot of sense to me, and I immediately accepted this explanation, but I did not write all the calculations down. Ten years later, I finally decided to write down how this works. So, not only it works, but we can also obtain the same effect in Squint using a Gaussian prior, that is a good enough approximation of the prior we implicitely used in 2016.

\section*{Acknowledgements}
I thank Haipeng Luo for discussions and feedback on this note.

\bibliographystyle{plainnat}
\bibliography{../../learning}

\begin{thebibliography}{4}
\providecommand{\natexlab}[1]{#1}
\providecommand{\url}[1]{\texttt{#1}}
\expandafter\ifx\csname urlstyle\endcsname\relax
  \providecommand{\doi}[1]{doi: #1}\else
  \providecommand{\doi}{doi: \begingroup \urlstyle{rm}\Url}\fi

\bibitem[Chernov and Vovk(2010)]{ChernovV10}
A.~Chernov and V.~Vovk.
\newblock Prediction with advice of unknown number of experts.
\newblock In \emph{Proc. of the Conference on Uncertainty in Artificial
  Intelligence (UAI)}, 2010.
\newblock URL \url{https://arxiv.org/abs/1408.2040}.

\bibitem[Koolen and van Erven(2015)]{KoolenVE15}
W.~M. Koolen and T.~van Erven.
\newblock Second-order quantile methods for experts and combinatorial games.
\newblock In \emph{Proc. of the Conference On Learning Theory}, pages
  1155--1175, 2015.
\newblock URL \url{https://arxiv.org/abs/1502.08009}.

\bibitem[Orabona(2019)]{Orabona19}
F.~Orabona.
\newblock A modern introduction to online learning.
\newblock \emph{arXiv preprint arXiv:1912.13213}, 2019.
\newblock URL \url{https://arxiv.org/abs/1912.13213}.
\newblock Version 9.

\bibitem[Orabona and P\'al(2016)]{OrabonaP16}
F.~Orabona and D.~P\'al.
\newblock Coin betting and parameter-free online learning.
\newblock In D.~D. Lee, M.~Sugiyama, U.~V. Luxburg, I.~Guyon, and R.~Garnett,
  editors, \emph{Advances in Neural Information Processing Systems 29}, pages
  577--585. Curran Associates, Inc., 2016.
\newblock URL \url{https://arxiv.org/pdf/1602.04128.pdf}.

\end{thebibliography}

\end{document}